\documentclass[runningheads]{llncs}

\usepackage{epsfig}
\usepackage{graphicx}
\usepackage{amsmath}
\usepackage{amssymb}
\usepackage{color}
\usepackage[width=122mm,left=12mm,paperwidth=146mm,height=193mm,top=12mm,paperheight=217mm]{geometry}
\begin{document}

\pagestyle{headings}
\mainmatter
\def\ECCV16SubNumber{***} 
%\def\cvprPaperID{1581} % *** Enter the CVPR Paper ID here
%\def\httilde{\mbox{\tt\raisebox{-.5ex}{\symbol{126}}}}

% Pages are numbered in submission mode, and unnumbered in camera-ready
%\ifcvprfinal\pagestyle{empty}\fi

%%%%%%%%% TITLE
%\title{Point-wise Mutual Information-based video segmentation with high temporal consistency}
\title{Point-wise mutual information-based video segmentation with high temporal consistency}
\author{Margret Keuper and Thomas Brox
% \and
% Thomas Brox\\
}
%\title{Author Guidelines for ECCV Submission} % Replace with your title

\titlerunning{Temporally consistent video segmentation}

\authorrunning{M. Keuper and T. Brox}

%\author{Anonymous ECCV submission}
\institute{Department of Computer science\\
University of Freiburg\\
Germany}

\maketitle
%\thispagestyle{empty}

%%%%%%%%% ABSTRACT
\begin{abstract}
In this paper, we tackle the problem of temporally consistent boundary detection and hierarchical segmentation in videos. While finding the best high-level reasoning of region assignments in videos is the focus of much recent research, temporal consistency in boundary detection has so far only rarely been tackled. We argue that temporally consistent boundaries are a key component to temporally consistent region assignment. The proposed method is based on the point-wise mutual information (PMI) of spatio-temporal voxels. Temporal consistency is established by an evaluation of PMI-based point affinities in the spectral domain over space and time. Thus, the proposed method is independent of any optical flow computation or previously learned motion models. The proposed low-level video segmentation method outperforms the learning-based state of the art in terms of standard region metrics.%   In this paper, we tackle the problem of temporally consistent boundary detection and hierarchical segmentation in videos. While finding the best high-level reasoning of region assignments in videos is the focus of much recent research, temporal consistency in boundary detection has so far only rarely been tackled. We argue however, that temporally consistent boundaries are a key component to temporally consistent region assignment. The proposed method is based on the point-wise mutual information (PMI) of spatio-temporal voxels. Temporal consistency is established by an evaluation of PMI-based point affinities in the spectral domain over space and time. Thus, the proposed method is independent of any optical flow computation and or previously learned motion models. Thereby, the proposed low-level video segmentation method outperforms the learning-based state of the art in terms of standard region metrics.    
\end{abstract}

%%%%%%%%% BODY TEXT
\section{Introduction}
 Accurate video segmentation is an important step in many high-level computer vision tasks. It can provide for example window proposals for object detection \cite{girshick15fastrcnn,ren2015faster} or action tubes for action recognition \cite{actiontubes,gkioxari2015wholesandparts}. One of the key challenges in video segmentation is on handling the large amount of data. Traditionally, methods either build upon some fine-grained image segmentation \cite{arbelaez-2011} or supervoxel \cite{corso1} method \cite{galasso-13,galasso-14,khoreva14,khoreva15,DBLP:journals/corr/YiP15} or they consist in the grouping of priorly computed point trajectories (e.g.\ \cite{Ochs14,keuper15b}) and transform them in a postprocessing step into dense segmentations \cite{OchsBroxICCV11}. The latter is well suited for motion segmentation applications, but has general issues with segmenting non-moving, or only slightly moving, objects.
Indeed, image segmentation into small segments forms the basis for many high-level video segmentation methods like \cite{galasso-14,khoreva15,DBLP:journals/corr/YiP15}. A key question when employing such preprocessing is the error it introduces. While state-of-the-art image segmentation methods \cite{arbelaez-2011,crisp_boundaries,APBMM2014} offer highly precise boundary localization, they usually suffer from low temporal consistency, i.e.,the superpixel shapes and sizes can change drastically from one frame to the next. This causes flickering effects in high-level segmentation methods.\\
\begin{figure}[t!]
\centering
\begin{tabular}{@{}l@{}l@{}l}
\includegraphics[width=0.333\linewidth]{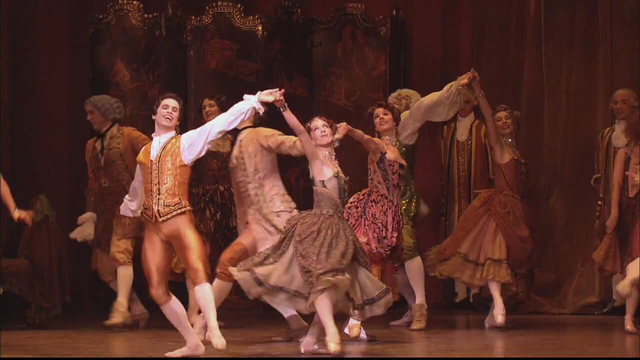}&
\includegraphics[width=0.333\linewidth]{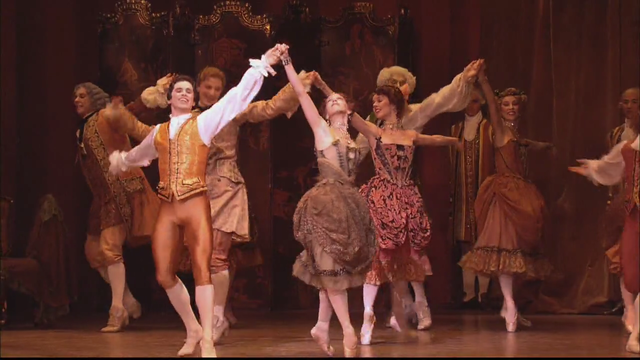}&
\includegraphics[width=0.333\linewidth]{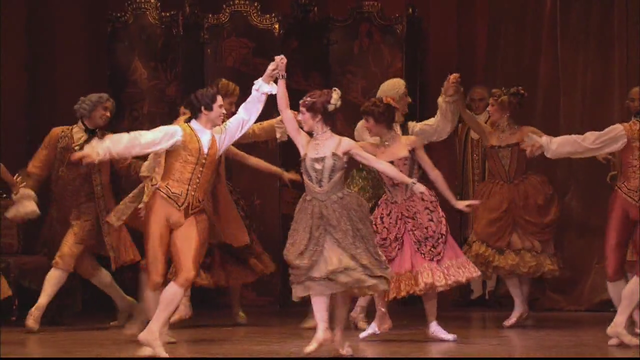}\\
\includegraphics[width=0.333\linewidth]{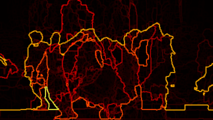}&
\includegraphics[width=0.333\linewidth]{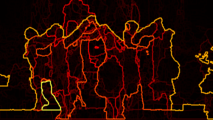}&
\includegraphics[width=0.333\linewidth]{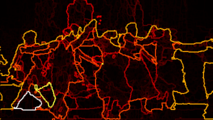}
\end{tabular}
\caption{Results of the proposed hierarchical video segmentation method for frame 4, 14 and 24 of the ballet sequence from VSB100 \cite{galasso-13}. The segmentation is displayed in a \emph{hot} color map. Note that corresponding contours have exactly the same value. Segmentations at different thresholds in this contour map are segmentations of the spatio-temporal volume.}%However, our curve remains in the low recall regime.}
\label{fig:teaser}
\end{figure}
%The wide use of superpixels and voxels for video segmentation suggests that advances in image segmentation are crucial for video segmentation.\\% Here, we propose to employ an affinity measure, that has recently been proposed for image segmentation \cite{crisp_boundaries}, in a video segmentation sceranio, in order to generate temporally consistent segmentations at different levels of coarsity.\\

In this paper, we present a low-level video segmentation method that aims at producing spatio-temporal superpixels with high temporal consistency in a bottom-up way. To this aim, we employ an affinity measure, that has recently been proposed for image segmentation \cite{crisp_boundaries}. While other, learning-based methods such as \cite{APBMM2014} slightly outperform \cite{crisp_boundaries} on the image segmentation task, they can hardly be transferred to video data because their boundary detection requires training data that is currently not available for videos. However, in \cite{crisp_boundaries}, boundary probabilities are learned in an unsupervised way from local image statistics, which can be transferred to video data.\\
To generate hierarchical segmentations, we build upon an established method for low-level image segmentation \cite{arbelaez-2011} and make it applicable to video data. More specifically, we build an affinity matrix according to \cite{crisp_boundaries} for each entire video over space and time.\\
Solving for the eigenvectors and eigenvalues of the resulting affinity matrices would require an enormous amount of computational resources. Instead, we show that solving the eigensystem for small temporal windows is sufficient and even produces results superior to those computed on the full system. To generate spatio-temporal segmentations from these eigenvectors according to what has been proposed in \cite{arbelaez-2011} for images, we need to generate small spatio-temporal segments from the eigenvectors. We do so by extending the oriented watershed transform \cite{arbelaez-2011} to three dimensions. This is substantial since apart from inferring object boundaries within a single frame it also allows us to predict where the same objects are located in the next frame. We achieve this without the need to compute optical flow. Instead, temporal consistency is maintained simply by the local affinities computed between frames and the smoothness within the resulting eigenvectors.\\
Once we have estimated the spatio-temporal boundary probabilities, we can apply the ultrametric contour map approach from \cite{ucm} on the three-dimensional data.\\
We show that the proposed low-level video segmentation method can compete with high-level learning-based approaches \cite{khoreva15} on the VSB100 \cite{galasso-13,Bro11c} video segmentation benchmark. In terms of temporal consistency, measured by the region metric VPR (volume precision recall), we outperform the state of the art.

%-------------------------------------------------------------------------
\subsection{Related work}  
\paragraph{Image Segmentation}
An important key to reliable image segmentation is the boundary detection. Most recent methods compute informative image boundaries with learning-based methods, either using random forests \cite{DollarICCV13edges,oriented} or convolutional neural networks \cite{gberta_2015_CVPR,DBLP:journals/corr/BertasiusST15,DBLP:journals/corr/XieT15}. Provided a sufficient amount of training data, these methods improve over spectral analysis-based methods \cite{arbelaez-2011,crisp_boundaries} that defined the state of the art before. However, the output of such methods provides a proxy for boundary probabilities but does not provide a segmentation into closed boundaries.\\ 
Actual segmentations can be built from these boundaries by the well-established oriented watershed and ultrametric contour map approach \cite{ucm} as in \cite{arbelaez-2011,crisp_boundaries,APBMM2014} or, as recently proposed, by minimum cost lifted multicuts \cite{keuper15a}.\\
Our proposed algorithm is most closely related to the image segmentation method from \cite{crisp_boundaries}, where the PMI-measure has been originally defined. The advantage of this measure is that is does not rely on any training data but estimates image affinities from local image statistics. We give some details of this approach in section \ref{sec:PMI}.
\paragraph{Video Segmentation}
The use of supervoxels and supervoxel hierarchies has been strongly promoted in recent video segmentation methods \cite{corso1,Chang2013tsp,corso3,GrundmannKwatra2010}. These supervoxels provide small spatio-temporal segments built from basic image cues such as color and edge information. While \cite{corso3} tackle the problem of finding the best supervoxel hierarchy flattening, \cite{GrundmannKwatra2010} build a graph upon supervoxels to introduce higher level knowledge. Similarly \cite{galasso-14,khoreva14,khoreva15,DBLP:journals/corr/YiP15} propose to build graphs upon superpixel segmentations and use learned \cite{khoreva14,khoreva15} information or multiple high-level cues \cite{DBLP:journals/corr/YiP15} to generate state-of-the-art video segmentations.\\
In \cite{galasso-12}, an attempt towards temporally consistent superpixels has been made on the bases of highly optimized image superpixels \cite{arbelaez-2011} and optical flow \cite{ldof}. Similar to the proposed method, \cite{galasso-12} try to make use of advances in image segmentation for video segmentation. However, their result still is a (temporally somewhat more consistent) frame-wise segmentation that is processed into a video segmentation by a graph-based method. In the benchmark paper \cite{galasso-13}, a baseline method for temporally consistent video segmentation has been proposed. From a state-of-the-art hierarchical image segmentation \cite{arbelaez-2011} computed on one video frame, the segmentation is propagated to the remaining frames by optical flow \cite{ldof}. The relatively good performance of this simple approach indicates that low-level cues from the individual video frames have high potential to improve video segmentation over the current state of the art.\\
In \cite{SundaramKeutzerICCV11} an extension of the method from \cite{arbelaez-2011} to video data has been proposed. In this work, the temporal link is established by optical flow \cite{ldof} and the pixel-wise eigensystem is solved for the whole video based on heavy gpu parallelization. Temporally consistent labelings are computed from the eigenvectors by direct spectral clustering, thus avoiding to handle the problem of temporal gradients.\\
In contrast, our method neither needs precomputed optical flow nor does it depend on solving the full eigensystem. Instead, temporal consistency is established by an in-between-frame evaluation of the point-wise mutual information \cite{crisp_boundaries}. Further, we extend the oriented watershed approach from \cite{arbelaez-2011} to the spatio-temporal domain so we can directly follow their approach in computing the ultametric contour map \cite{ucm}. In this setup, we can show that solving the eigensystem for temporal windows even improves over segmentations computed from solving the full eigensystem.
\section{Method Overview}
\begin{figure}[t!]
\centering
\includegraphics[width=0.45\linewidth]{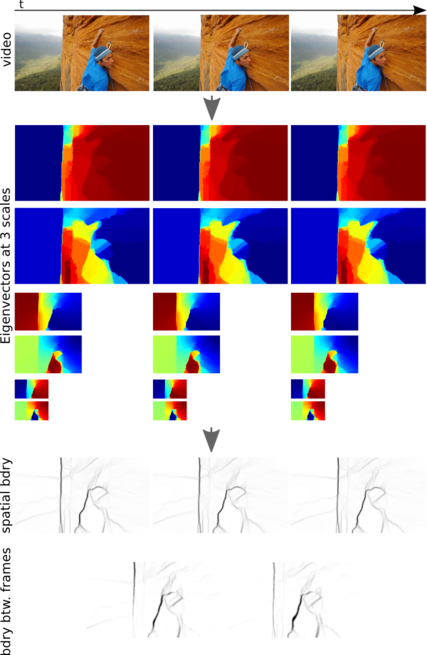} \includegraphics[width=0.45\linewidth]{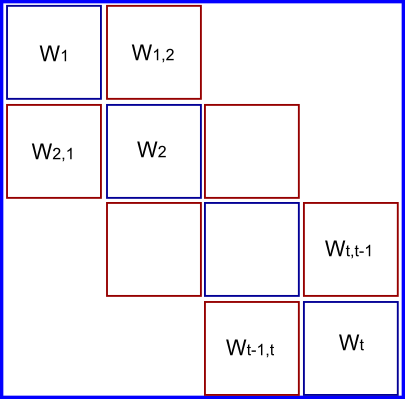}
\caption{Workflow of the proposed method for temporally consistent boundary detection. From an image sequence, temporally consistent eigenvectors are computed at multiple scales. While the spatial boundaries indicate object boundaries for every frame, boundaries in-between frames indicate the changes over time. In the example sequence, the \emph{rock climbing} sequence from the VSB100 \cite{galasso-13,Bro11c} training set, the dominant change over time in the first frames is the camera motion.}
\label{fig:workflow}
\end{figure}

The proposed method is a video segmentation adaptation of a pipeline that has been used in several previous works on image segmentation \cite{arbelaez-2011,crisp_boundaries,APBMM2014} with slight variations. The key steps are given in Fig. \ref{fig:workflow}. We start from the entire video sequence and compute a full affinity matrix at multiple scales. Eigenvectors are computed for these scales within small overlapping temporal windows of three frames. On the three-dimensional spatio-temporal volumes of eigenvectors, spatial and temporal boundaries can be estimated. These can be fed into the ultrametric contour map hierarchical segmentation \cite{ucm} adapted for three-dimensional data. 
\section{Point-wise mutual information}
\label{sec:PMI}
We follow \cite{crisp_boundaries} in defining the point-wise mutual information (PMI) measure employed for the definition of pairwise affinities. Let the random variables $A$ and $B$ denote a pair of neighboring features. In \cite{crisp_boundaries}, the joint probability $P(A,B)$ is defined as a weighted sum of the joint probability $p(A,B;d)$ of features $A$ and $B$ occurring with a Euclidean distance of $d$:
\begin{equation}
P(A,B)=\frac{1}{Z}\sum_{d=d_0}^\infty w(d)p(A,B;d).
\end{equation}
Here,$Z$ is a normalization constant and the weighting function $w$ is a Gaussian normal distribution with mean-value two. The marginals of the above distribution are used to define $P(A)$ and $P(B)$. To define the affinity of two neighboring points, the direct use of this the joint probability has the disadvantage of being biased by the frequency of occurrence of $A$ and $B$, i.e.\ if a feature occurs frequently in an image, the feature will have a relatively high probability to co-occur with any other feature. The PMI corrects for this unbalancing:
\begin{equation}
\text{PMI}_\rho(A,B)=\text{log}\frac{P(A,B)^\rho}{P(A)P(B)}.
\end{equation}
In \cite{crisp_boundaries}, the parameter $\rho$ is optimized on the training set of the BSD500 image segmentation benchmark \cite{arbelaez-2011}. We stick to their resulting parameter choice of $\rho=1.25$.\\
The crucial part of the affinity measure from \cite{crisp_boundaries} that makes it easily applicable to unsupervised boundary detection is that $P(A,B)$ is learned specifically for every image from local image statistics. More specifically, for 10000 random sample locations per image, features $A$ and $B$ with mutual distance $d$ are sampled. To model the distribution, kernel density estimation \cite{kde} is employed.\\
\subsection{Spatio-temporal affinities}
\label{sec:affinity}
According statistics could be computed on entire videos and used to generate segmentations. However, there is a strong reason not to compute the $P(A,B)$ over all video frames: image statistics can change drastically during a video or image sequence, e.g.\ when new agents enter the scene, the camera moves or the illumination changes. To be robust towards these changes, we chose to estimate $P(A,B)$ per video frame and use these for the computation of affinities within this frame within this frame and in-between this frame and the next. This is justified by the assumption that changes in local statistics are temporally smooth.\\
Thus, within every frame $t$, affinities of its elements are computed according to the estimated $\text{PMI}_{\rho,t}$ for color and local variance of every pixel to every pixel within a radius of $5$ pixels. Within every frame $t$ and its successor $t+1$, affinities are computed according to $\text{PMI}_{\rho,t}$ for every pixel in frame $t$ to every pixel in frame $t+1$ within spatial distance of $3$ pixels, and, for every pixel in frame $t+1$ to every pixel in frame $t$ within the same distance. The result is a sparse symmetric spatio-temporal affinity matrix $W$ as given in Fig. \ref{fig:workflow}. 

% \begin{figure}[t]
% \begin{center}
% %\fbox{\rule{0pt}{2in} \rule{0.9\linewidth}{0pt}}
%    \includegraphics[width=0.3\linewidth]{affinitysym.png}
% \end{center}
%    \caption{Spatio-temporal affinity matrix. The within-frame affinity matrices $W_i$ are located on the diagonal of $W$ to build a block-diagonal matrix. The in-between-frame affinity matrices are located around the diagonal. Since we consider only affinities between neighboring frames, the resulting $W$ is very sparse.}
% \label{fig:affinity}
% \end{figure}
\section{Spectral Boundary Detection}
\begin{figure}[t!]
\centering
\begin{tabular}{@{}l@{}l@{}l@{}}
 \includegraphics[width=0.266\linewidth]{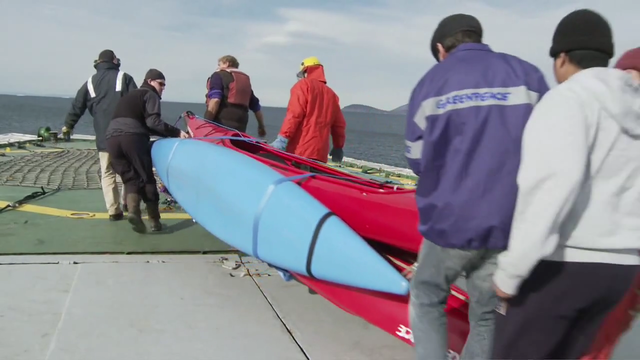}&
\includegraphics[width=0.266\linewidth]{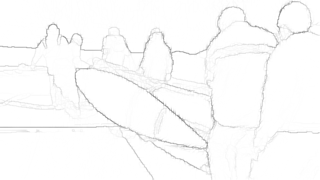}&
\includegraphics[width=0.266\linewidth]{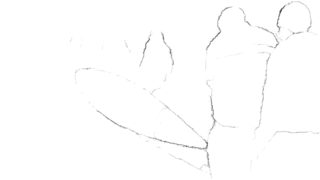}\\
\includegraphics[width=0.266\linewidth]{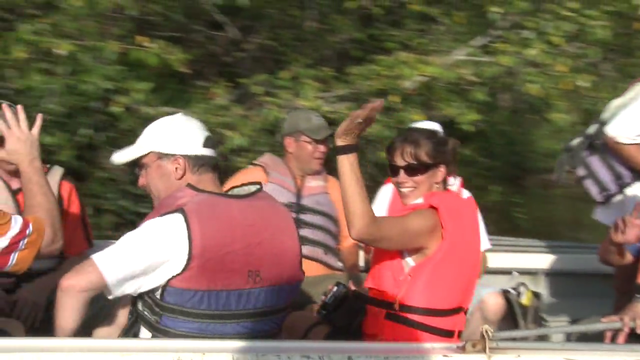}&
\includegraphics[width=0.266\linewidth]{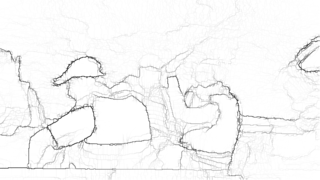}&
\includegraphics[width=0.266\linewidth]{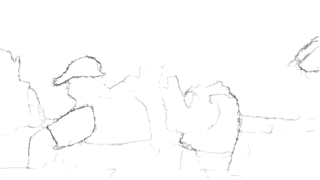}\\
\includegraphics[width=0.266\linewidth]{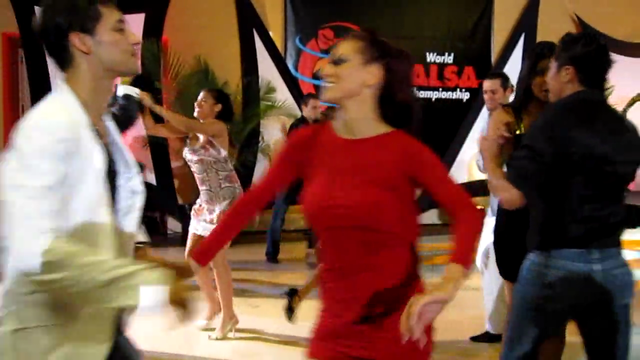}&
\includegraphics[width=0.266\linewidth]{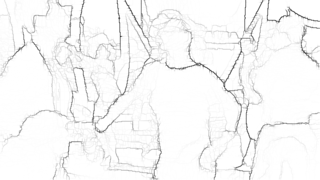}&
\includegraphics[width=0.266\linewidth]{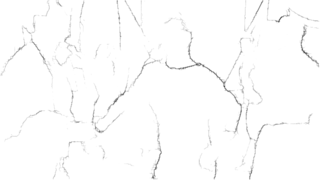}\\
\includegraphics[width=0.266\linewidth]{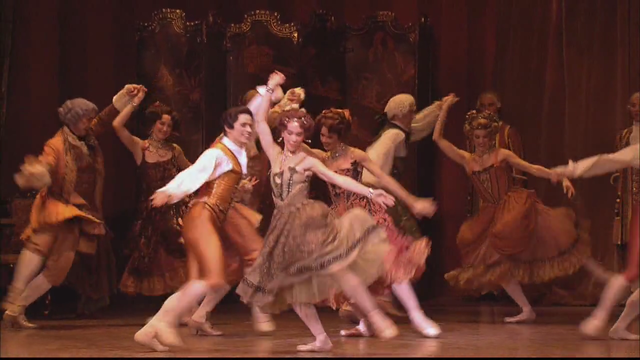}&
\includegraphics[width=0.266\linewidth]{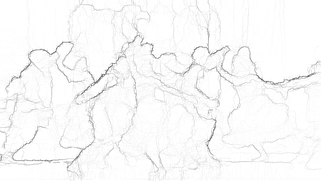}&
\includegraphics[width=0.266\linewidth]{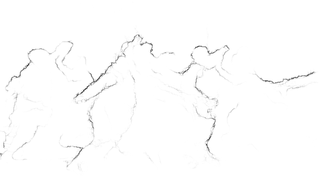}\\
image&all bdry& motion bdry 
\end{tabular}
\caption{Examples of boundary estimates (center) and motion boundary estimates (right) on three sequences of VSB100 \cite{galasso-13,Bro11c} (\emph{arctic kayak}, \emph{riverboat}, and \emph{salsa}). The motion boundaries can be directly derived from the proposed method.}%However, our curve remains in the low recall regime.}
\label{fig:boundaries}
\end{figure}

Given an affinity matrix $W$, spectral clustering can be employed to generate boundary probabilities \cite{arbelaez-2011,crisp_boundaries} and segmentations \cite{galasso-12,galasso-14,khoreva15} according to a balancing criterion, more precisely, approximating the normalized cut
\begin{equation}\label{eq:thncut}
\text{NCut}(A,B)=\frac{\text{cut}(A,B)}{\text{vol}(A)}+\frac{\text{cut}(A,B)}{\text{vol}(B)},
\end{equation}
with $\;\displaystyle\text{cut}(A,B)= \hspace{-0.08cm}\sum_{i\in A, j\in B}\hspace{-0.06cm}w_{ij}\;$ and $\;\displaystyle\text{vol}(A)=\hspace{-0.08cm}\sum_{i\in A,j\in\mathcal{V}}\hspace{-0.06cm} w_{ij}$.\\
Approximate solutions to the normalized cut are induced by the first $k$ eigenvectors of the normalized graph Laplacian $L_{\text{sym}}=I-D^{-\frac{1}{2}}WD^{-\frac{1}{2}}$, where $D$ is the diagonal degree matrix of $W$ computed by $d_{ii}=\sum_{j\in{V}}w_{ij}$. \\
However, the computation of eigenvectors for large affinity matrices rapidly becomes expensive both in terms of computation time and memory consumption. To keep the computation tractable, we can reduce the computation to small temporal windows and employ the spectral graph reduction \cite{galasso-14} technique.\\
\paragraph{Spectral Graph Reduction}
Spectral graph reduction \cite{galasso-14} is a means of solving a spectral clustering or normalized cut problem on a reduced set of points. In this setup, the matrix $W$ defines the edge weights in a graph $G=(V,E)$. Given some pre-grouping of vertices by for example superpixels or must-link constraints, \cite{galasso-14} specify how to set these weights in a new graph $G' = (V',E')$ where $V'$ represents the set of vertex groups and $E'$ the set of edges in between them such that the normalized cut objective does not change. They show on low-level image segmentation as well as on high-level video segmentation the advantages of this method.\\
Since we want to remain as close to the low-level problem as possible, we employ a setup similar to the one proposed in \cite{galasso-14} for image segmentation. More specifically, we compute superpixel at the finest level produced by \cite{APBMM2014} for every frame, which builds upon learned boundary probabilities from \cite{DollarICCV13edges}. In order not to lose accuracy in boundary localization, \cite{galasso-14} proposed to keep single pixels in all regions with high gradients and investigate the trade-off between pixels and superpixels that is necessary. Similarly, we keep single pixels in all regions with high boundary probability.\\
%TODO: resulting graph sizes for videos
\paragraph{Multiscale Approach and Boundary Detection}
Since it has been shown in the past that spectral clustering based methods benefit from multi-scale information, we build affinity matrices also for videos spatially downsampled by factor 2 and factor 4. In this case, no pixel pre-grouping is necessary. For all three scales, we solve the eigensystems individually and compute the smallest 20 eigenvalues and according eigenvectors (compare Fig. \ref{fig:workflow}). Note that these eigenvectors are highly consistent over the temporal dimension. We upsample these vectors to the highest resolution and compute oriented edges in $9$ directions, with the standard oriented edge filters in 8  sampled spatial orientations and only one temporal gradient, i.e.\ there is no mixed spatio-temporal gradient. Depending on the frame-rate, using finer orientation sampling would certainly make sense. However, on the VSB100 dataset \cite{galasso-13,Bro11c}, we found that this simple setup works best. Examples of our extracted boundary estimates are given in Fig. \ref{fig:boundaries}. Visually, the estimated boundaries look reasonable and are temporally highly consistent. They form the key to the final hierarchical segmentations.
%\paragraph{Evaluation of Within-Frame Boundaries}
%Visually, the estimated boundaries look reasonable and are temporally highly consistent. They form the key to the final hierarchical segmentations. To assess their quality in terms of frame-wise boundary localization, we can measure the boundary precision and recall of the (thinned) boundary probabilities on the VSB100 video segmentation benchmark \cite{galasso-13} which consists of 40 train and 60 test sequences with a maximum length of 121 frames. Human segmentation annotations are given for every 20th frame.
\paragraph{Evaluation of Temporal Boundaries} On the BVSD \cite{Bro11c} dataset, benchmark annotations for occlusion boundaries were provided. This data can be used as a proxy to evaluate our temporal boundaries. Occlusion boundaries are object boundaries that occlude other parts of the scene - as opposed to within-object boundaries. In \cite{Bro11c}, the importance of motion cues for such occlusion boundaries has been pointed out. In fact, our temporal boundaries indicate boundaries separating regions within one frame that will undergo occlusion or disocclusion between this frame and the next. Thus, they can only provide part of the necessary information for object boundary detection. To extract this motion cue from our data, we apply a pointwise multiplication of the spatial boundaries at frame $t$ with the temporal boundaries between frame $t$ and its two neigh boring frames. Thus, if an object does not move in one of the frames, the respective edges are removed. Examples of the resulting motion boundary estimates are given in Fig. \ref{fig:boundaries}.\\
 When we evaluate on the closed boundary annotations of this benchmark with the benchmark parameters from BSDS500 \cite{arbelaez-2011}, we get a surprisingly low f-measure score of 0.34 with the best common threshold for the whole dataset, 0.41 if we allow individual thresholds per sequence. The reason might be the relatively low spatial localization accuracy of our boundaries. We ran all our experiments on the \emph{half resolution} version of the VSB100 benchmark such that, to evaluate on the annotations from  \cite{Bro11c}, boundary estimates need to be upsampled.
%-------------------------------------------------------------------------
\begin{figure*}[t!]
\centering
\begin{tabular}{@{}c@{}c@{}c}
\multicolumn{3}{c}{\includegraphics[width=0.79\linewidth]{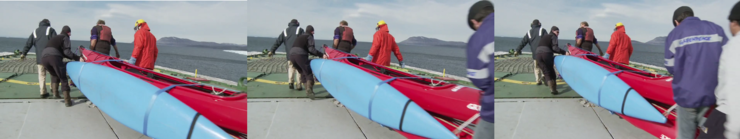}}\\
\multicolumn{3}{c}{\includegraphics[width=0.79\linewidth]{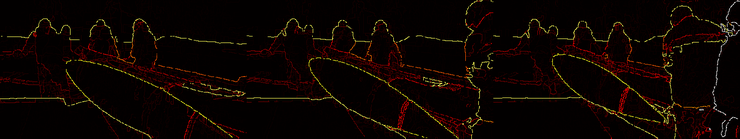}}\\
\multicolumn{3}{c}{\includegraphics[width=0.79\linewidth]{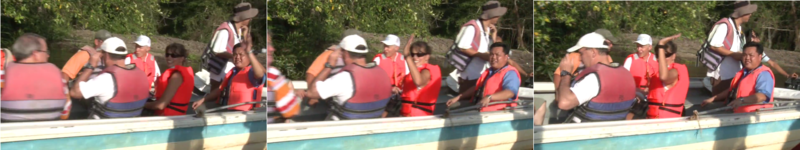}}\\
\multicolumn{3}{c}{\includegraphics[width=0.79\linewidth]{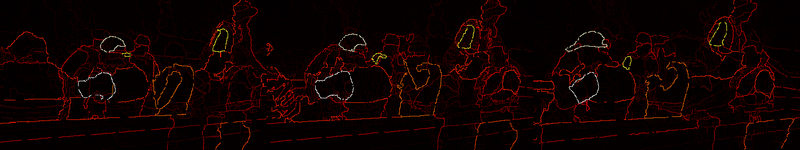}}\\
%\footnotesize{frame 1}&\footnotesize{frame 11}&\footnotesize{frame 21}\\
\end{tabular}
\caption{Resulting UCMs from the proposed method on the sequences \emph{arctic kayak} and \emph{riverboat} of the VSB100 \cite{galasso-13,Bro11c} dataset. We show the UCM value in frames 1, 11, and 21 while these actually extend over the full lengths of the sequences. Especially in the high levels of the hierarchy, the segmentations are very consistent.}%However, our curve remains in the low recall regime.}
\label{fig:ucms}
\end{figure*}
\section{Closed Spatio-temporal Contours}
Given spatio-temporal boundary estimates, closed regions could be generated by different methods such as region growing, agglomerative clustering \cite{gw}, watersheds \cite{watershed} or the recently proposed minimum cost lifted multicuts \cite{keuper15a}. In \cite{ucm} a mathematically sound and widely used (e.g.\ in \cite{arbelaez-2011,SundaramKeutzerICCV11,galasso-12,crisp_boundaries,APBMM2014}) setup for the generation of hierarchical segmentations from boundary probabilities and an initial fine-grained segmentation is given. The therein defined Ultrametric Contour Map provides for a duality between the saliency of a contour and the scale of its disappearance from the hierarchy.\\
The approach from \cite{ucm}  can directly be applied to three-dimensional data. The difference is that the region contours are now two-dimensional curves that meet each other in one-dimensional curves or points. Each one-dimensional curve is common to at least three contours. As in the two-dimensional case, every contour is separating exactly two regions.\\
Samples from the resulting Ultrametric Contour Maps can be seen in Fig. \ref{fig:ucms}. The brightness of the contour, displayed in \emph{hot} color maps, indicate the saliency of a contour, i.e.\ its hierarchical level in the segmentation. Over all frames of the videos, the resulting closed contours have consistent saliency.
%Definition from \cite{ucm}: the union over two regions $R1$ and $R2$ must have distances to its adjacent regions larger or equal to the distances of either $R1$ or $R2$.\\
%Spectral clustering for spatial consistency in sequences with strong motion.\\
%Acts as must-link constraints over time.
%\section{Relationship to Lifted Multicuts}
%+ No balancing criterion.\\
%+ Output is an optimal partition of the spatio-temporal volume into an optimal number of segments.\\
%- The LMP is hard to optimize - the minimum cost multicut problem is APX-hard. In \cite{keuper15a}, a primal heuristic has been used to find approximate solutions that do not provide any lower bounds on the approximation quality but perform well in praxis.
%- There is no direct translation into interpixel boundary probabilities from the lifted multicut problem.
\section{Experiments and Results}
\paragraph{Setup}
We compute PMI-based affinity matrices $W_s$ on color and local variance within all frames and between every frame and its successive frame as described in section \ref{sec:affinity} for three different scales (1, 0.5 and 0.25). For scale 1, we employ spectral graph reduction \cite{galasso-14}, reducing the number of nodes by factor 12-15.
At each scale $s$, we solve the eigenvalue problems of the normalized graph Laplacians corresponding to $W_s$ for overlapping temporal windows with stride 1 to generate the first 20 eigenvectors. The best choice of the temporal window size is not obvious because of the eigenvector leakage problem, also mentioned in \cite{SundaramKeutzerICCV11}. In the spectrally reduced graph, spatial leakage is probably low \cite{galasso-14}, so we we hope for an accordingly low temporal leakage and choose a larger temporal window of size 5, while we solve the eigenvalue problem for smaller temporal windows of length 3 for scales 0.5 and 0.25. The resulting eigenvectors are resampled to the original resolution. The average oriented gradients on these eigenvectors for the multiscale boundary estimates we use. We compare the 3D ultrametric contour maps computed from these boundary estimates to those computed on only the original scale with spectral graph reduction. For the original scale, we also compare to the results we get by solving the eigensystem on the full video without temporal windows.
\begin{figure*}[t!]
\centering
\begin{tabular}{@{}c@{}c@{}}
%\multicolumn{2}{c}{\textbf{Motion Segmentation} }& \multicolumn{2}{c}{\textbf{Non-rigid Motion} }\\
\includegraphics[width=0.49\linewidth]{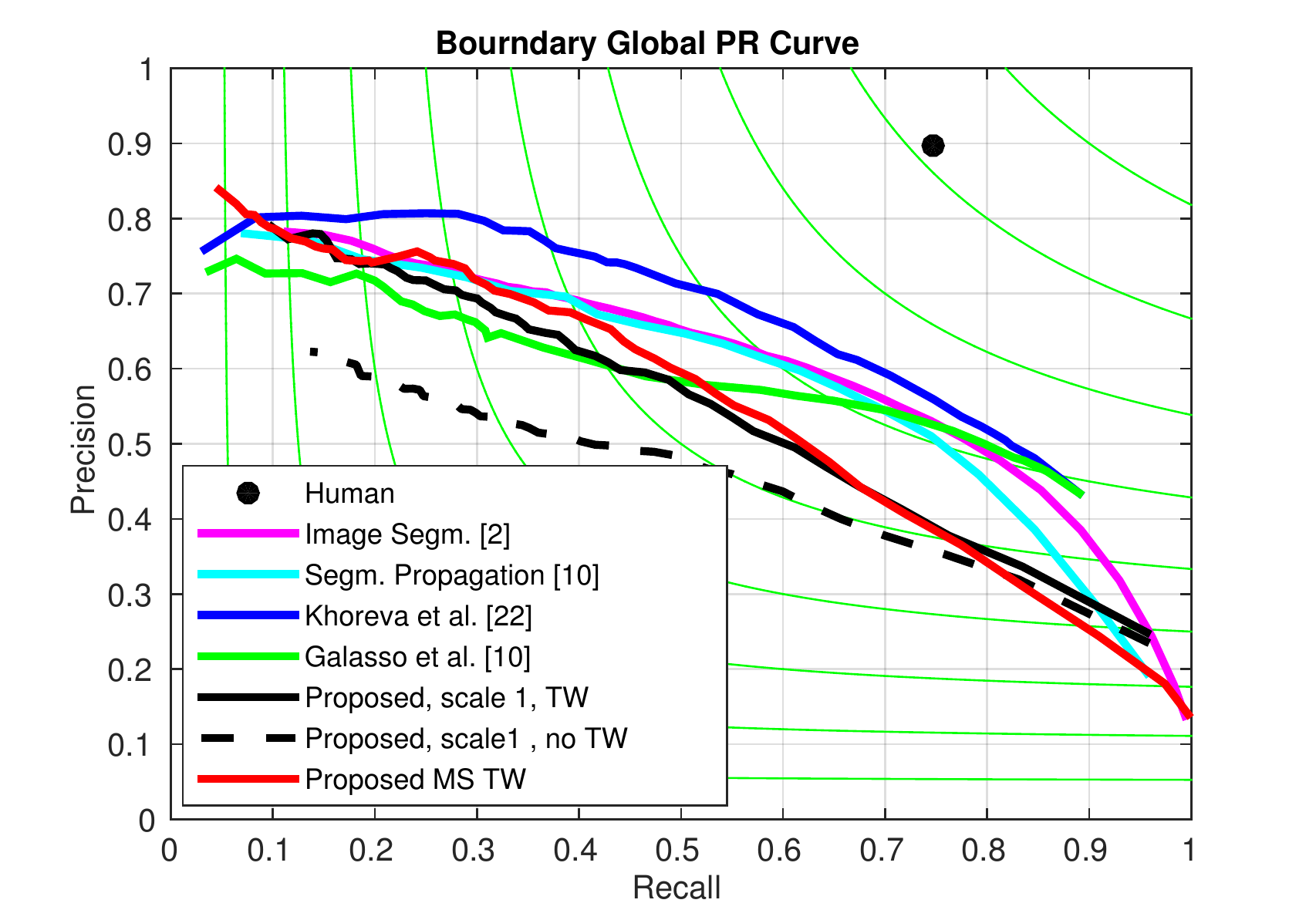}&
\includegraphics[width=0.49\linewidth]{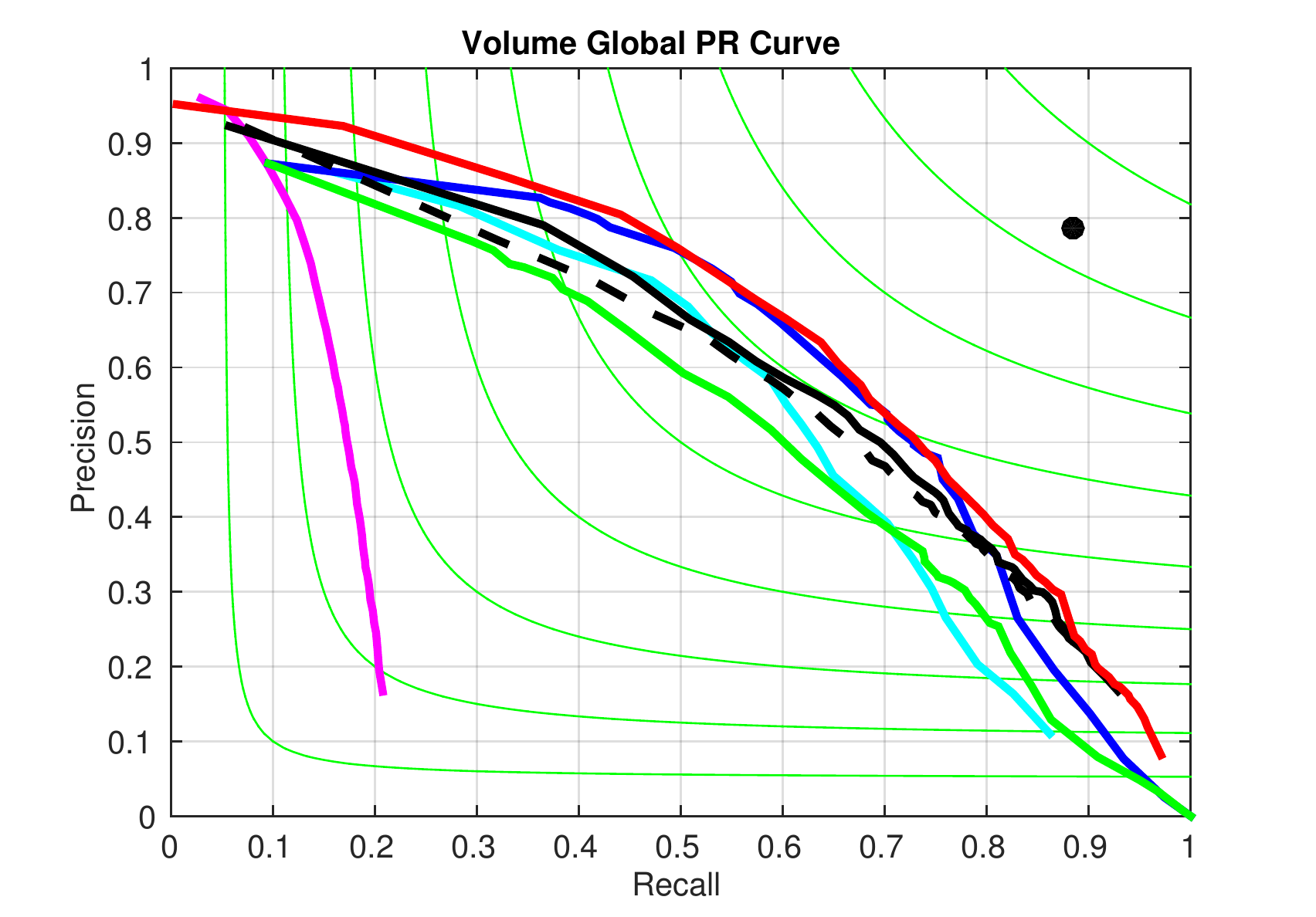}\vspace{-0.1cm}\\
\footnotesize{BPR}&\footnotesize{VPR}\\
\end{tabular}
\caption{Results on the general benchmark of the VSB100 \cite{galasso-13,Bro11c} dataset. While the proposed method performs worse than the state of the art in terms of boundary precision and recall (BPR), we outperform all competing methods on the region metric VPR.}%However, our curve remains in the low recall regime.}
\label{fig:VSB100_1}
\end{figure*} 
\begin{table*}[t!]
\begin{center}\begin{small}
\begin{tabular}{l|ccc|ccc}
\multicolumn{7}{c}{VSB100: general benchmark}\\
 & \multicolumn{3}{c|}{BPR} & \multicolumn{3}{c}{VPR} \\
\hline Algorithm & \;ODS\; & \;OSS\; & \;\;AP\;\; & \;ODS\; & \;OSS\; & \;\;AP\;\; \\
\hline Human
    & 0.71 & 0.71 & 0.53 & 0.83 & 0.83 & 0.70 \\
%\hline $^{*}$Corso et al. \cite{CorsoetalTMI08}
   % & 0.51 & 0.53 & 0.37 & 0.51 & 0.52 & 0.38 & 70.67(48.39) & 25.83\\
\hline $^{*}$Galasso et al. \cite{galasso-12}
    & 0.52 & 0.56 & 0.44 & 0.45 & 0.51 & 0.42 \\ 
 $^{*}$Grundmann et al. \cite{GrundmannKwatra2010}
    & 0.47 & 0.54 & 0.42 & 0.52 & 0.55 & 0.52 \\
 $^{*}$Ochs and Brox \cite{OchsBroxICCV11}
    & 0.14 & 0.14 & 0.04 & 0.25 & 0.25 & 0.12 \\
Xu et al. \cite{corso2}
    & 0.40 & 0.48 & 0.33 & 0.45 & 0.48 & 0.44 \\
%\hline $^{*}\mathcal{G}''$ (SPX Lev.2) & 0.61 & \textbf{0.65} & 0.48 & 0.44 & 0.50 & 0.42 & 60.42(41.85) & 18.00\\
%\hline $^{*}$ $\mathcal{G}^Q\equiv\mathcal{G}'$ Ncut & \textbf{0.62} & \textbf{0.65} & \textbf{0.50} & 0.54 & 0.57 & 0.52 & 69.08(40.91) & 20.00\\
%\hline $^{*}$ $\mathcal{G}^Q\equiv\mathcal{G}^0$ Ncut & \textbf{0.62} & \textbf{0.65} & 0.49 & \textbf{0.55} & \textbf{0.59} & 0.53 & 71.77(40.27) & 20.00\\
$^{*}$ Galasso et al. \cite{galasso-14} & {0.62} & {0.65} & {0.50} & {0.55} & {0.59} & {0.55} \\
%\hline $^{*}$ $\mathcal{G}^Q\equiv\mathcal{G}^0$ SC & \textbf{0.62} & \textbf{0.65} & 0.49 & 0.53 & 0.57 & 0.52 & 52.81(46.25) & 100.00\\\
 $^{*}$ Khoreva et al. \cite{khoreva15} SC & \textbf{0.64} & \textbf{0.70} & \textbf{0.61} & 0.63 & \textbf{0.66} & 0.63\\
%\hline $^{*}$ $\mathcal{G}^Q\equiv\mathcal{G}^0$ SC & \textbf{0.62} & \textbf{0.65} & 0.49 & 0.53 & 0.57 & 0.52 & 52.81(46.25) & 100.00\\

\hline Segmentation Propagation \cite{galasso-13}   %thr 0, nthr 20
    &0.60 &0.64& 0.57 & {0.59} &{0.62} &{0.56}\\
 IS - Arbelaez et al. \cite{arbelaez-2011}
    & 0.61 & 0.65 & 0.61 & 0.26 & 0.27 & 0.16 \\
 Oracle \& IS - Arbelaez et al. \cite{arbelaez-2011}
    & 0.61 & 0.67 & 0.61 & 0.65 & 0.67 & 0.68  \\ %1.00(0.00)

\hline $^{*}$ Proposed MS TW   %thr 0, nthr 20
    & 0.56 & 0.63 & {0.56} & \textbf{0.64} & \textbf{0.66} & \textbf{0.67} \\%6.1(9.41) & 2275.32\\
\end{tabular}
\end{small}\end{center}
\caption{Comparison of state-of-the-art video segmentation algorithms~\cite{GrundmannKwatra2010,OchsBroxICCV11,galasso-12,corso2,galasso-14,khoreva15,galasso-13} with the proposed low-level method. It shows the boundary precision-recall (BPR) and volume precision-recall (VPR) at optimal dataset scale (ODS) and optimal segmentation scale (OSS), as well as the average precision (AP). The algorithms marked with (*) have been evaluated on resized video frames to 0.5 of the original resolution.
% We report the additional statistics of mean ($\mu$) and standard deviation ($\delta$) volume lengths [Length] and number of clusters [NCL]. (*) indicates evaluated on video frames resized by 0.5 in the spatial dimension.
}
%\vspace{-0.5cm}
\label{tab:vs_bvds_gen}
\end{table*}   
\paragraph{Evaluation}
We evaluate the proposed video segmentation method on the \emph{half resolution} version of the VSB100 video segmentation benchmark \cite{galasso-13,Bro11c}. It consists of 40 train and 60 test sequences with a maximum length of 121 frames. Human segmentation annotations are given for every 20th frame. Two evaluation metrics are relevant in this benchmark, denoted as boundary precision and recall (BPR) and the volume precision recall (VPR). The BPR measures the accuracy of the boundary localizations per frame. Image segmentation methods usually perform well on this measure, since temporal consistency is not taken into account. The VPR is a region metric. Here, exact boundary localization is less important, while the focus lies on the temporal consistency. This is the measure on which we expect to perform well.\\        
\begin{table*}[t!]
\begin{center}\begin{small}
\begin{tabular}{l|ccc|ccc}
\multicolumn{7}{c}{VSB100: motion subtask}\\
 & \multicolumn{3}{c|}{BPR} & \multicolumn{3}{c}{VPR} \\
\hline Algorithm & \;ODS\; & \;OSS\; & \;\;AP\;\; & \;ODS\; & \;OSS\; & \;\;AP\;\; \\
\hline Human
    & 0.63 & 0.63 & 0.44 & 0.76 & 0.76 & 0.59 \\
%\hline $^{*}$Corso et al. \cite{CorsoetalTMI08}
   % & 0.51 & 0.53 & 0.37 & 0.51 & 0.52 & 0.38 & 70.67(48.39) & 25.83\\
\hline $^{*}$Galasso et al. \cite{galasso-12}
    & 0.34 & 0.43 & 0.23 & 0.42 & 0.46 & 0.36 \\ 
 $^{*}$Ochs and Brox \cite{OchsBroxICCV11}
    & 0.26 & 0.26 & 0.08 & 0.41 & 0.41 & 0.23 \\
%Xu et al. \cite{corso2}
%    & 0.40 & 0.48 & 0.33 & 0.45 & 0.48 & 0.44 \\
%\hline $^{*}\mathcal{G}''$ (SPX Lev.2) & 0.61 & \textbf{0.65} & 0.48 & 0.44 & 0.50 & 0.42 & 60.42(41.85) & 18.00\\
%\hline $^{*}$ $\mathcal{G}^Q\equiv\mathcal{G}'$ Ncut & \textbf{0.62} & \textbf{0.65} & \textbf{0.50} & 0.54 & 0.57 & 0.52 & 69.08(40.91) & 20.00\\
%\hline $^{*}$ $\mathcal{G}^Q\equiv\mathcal{G}^0$ Ncut & \textbf{0.62} & \textbf{0.65} & 0.49 & \textbf{0.55} & \textbf{0.59} & 0.53 & 71.77(40.27) & 20.00\\
%$^{*}$ Galasso et al. \cite{galasso-14} & {0.62} & {0.65} & {0.50} & {0.55} & {0.59} & {0.55} \\
%\hline $^{*}$ $\mathcal{G}^Q\equiv\mathcal{G}^0$ SC & \textbf{0.62} & \textbf{0.65} & 0.49 & 0.53 & 0.57 & 0.52 & 52.81(46.25) & 100.00\\\
 $^{*}$ Khoreva et al. \cite{khoreva15}  & 0.45 & \textbf{0.53} & 0.33 & 0.56 & \textbf{0.63} & 0.56\\
%\hline $^{*}$ $\mathcal{G}^Q\equiv\mathcal{G}^0$ SC & \textbf{0.62} & \textbf{0.65} & 0.49 & 0.53 & 0.57 & 0.52 & 52.81(46.25) & 100.00\\

\hline Segmentation Propagation \cite{galasso-13}   %thr 0, nthr 20
    &\textbf{0.47} & 0.52 & \textbf{0.34} & {0.52} &{0.57} &{0.47}\\
 IS - Arbelaez et al. \cite{arbelaez-2011}
    & 0.47 & 0.43 & 0.35 & 0.22 & 0.22 & 0.13 \\
 Oracle \& IS - Arbelaez et al. \cite{arbelaez-2011}
    & 0.47 & 0.34 & 0.35 & 0.59 & 0.60 & 0.60  \\ %1.00(0.00)

\hline $^{*}$ Proposed MS TW   %thr 0, nthr 20
    & 0.41 & 0.43 & {0.29} & \textbf{0.58} & 0.58 & \textbf{0.58} \\%6.1(9.41) & 2275.32\\
\end{tabular}
\end{small}\end{center}
\caption{Comparison of state-of-the-art video segmentation algorithms with the proposed low-level method on the motion subtask of VSB100 \cite{galasso-13}.
% We report the additional statistics of mean ($\mu$) and standard deviation ($\delta$) volume lengths [Length] and number of clusters [NCL]. (*) indicates evaluated on video frames resized by 0.5 in the spatial dimension.
}
%\vspace{-0.5cm}
\label{tab:vs_bvds_motion}
\end{table*}
\paragraph{Results}
The results of our PMI-based video segmentation are given in Fig. \ref{fig:VSB100_1} in terms of BRP and VPR curves for 51 different levels of segmentation granularity, which is the standard for the VSB100 video segmentation benchmark.  In terms of BPR, all our results remain below the state of the art. There can be several reasons for this behavior: (1) Competitive methods that are based on per image superpixels usually compute these superpixels on the highest possible resolution \cite{galasso-14,khoreva15}, while we start from the \emph{half resolution} version of the benchmark data. (2) The number of eigenvectors we compute might be too low. In our current setup, we compute the first 20 eigenvectors per matrix. The optimal number to be chosen here varies strongly depending on the structure of the data as well as the employed affinities. (3) Most importantly, by the definition of our boundary detection method, we force these boundaries to be spatially consistent. However, spatial consistency is not at all required for this measure. If our boundaries show some amount of temporal smoothness, this might cause the boundaries to be slightly shifted in an individual frame.\\
On the VPR, the proposed method benefits from the temporal consistency it optimizes and outperforms all previous methods in the three aggregate measures ODS (meaning, we choose one global segmentation threshold for the whole dataset), in OSS, which allows to choose the best threshold per sequence, and average precision (AP). Respective numbers are given in Tab. \ref{tab:vs_bvds_gen}. This gives rise to the conclusion that the proposed segmentations are indeed temporally consistent.\\
In Fig. \ref{fig:VSB100_1}, we also plot the result we get if we only use boundary estimates from the original resolution without using multiscale information (depicted in black). As we expect, the segmentation quality remains below the quality we get with the multiscale approach. However, we did not know a priori what to expect from solving the eigensystem for the entire videos without applying temporal windows (dashed black line). To do so actually requires large amounts of memory ($>512$Gb) for most sequences. The results actually remain clearly below those computed with temporal windows. In fact, the eigenvectors we compute on the small temporal windows show high temporal consistency while those computed on the whole video are subject to temporal leakage of the eigenvectors, meaning that values in the eigenvectors within a region can change smoothly throughout the sequence, resulting in decreased discrimination power.\\
The best results in Tab. \ref{tab:vs_bvds_gen} are those from the oracle case, where the individual image segments from \cite{arbelaez-2011} are temporally linked based on the ground truth. The numbers indicate the best result one could achieve on the benchmark, starting from the given image segmentation.\\
\begin{figure}[t!]
\centering
%\multicolumn{2}{c}{\textbf{Motion Segmentation} }& \multicolumn{2}{c}{\textbf{Non-rigid Motion} }\\
\includegraphics[width=0.49\linewidth]{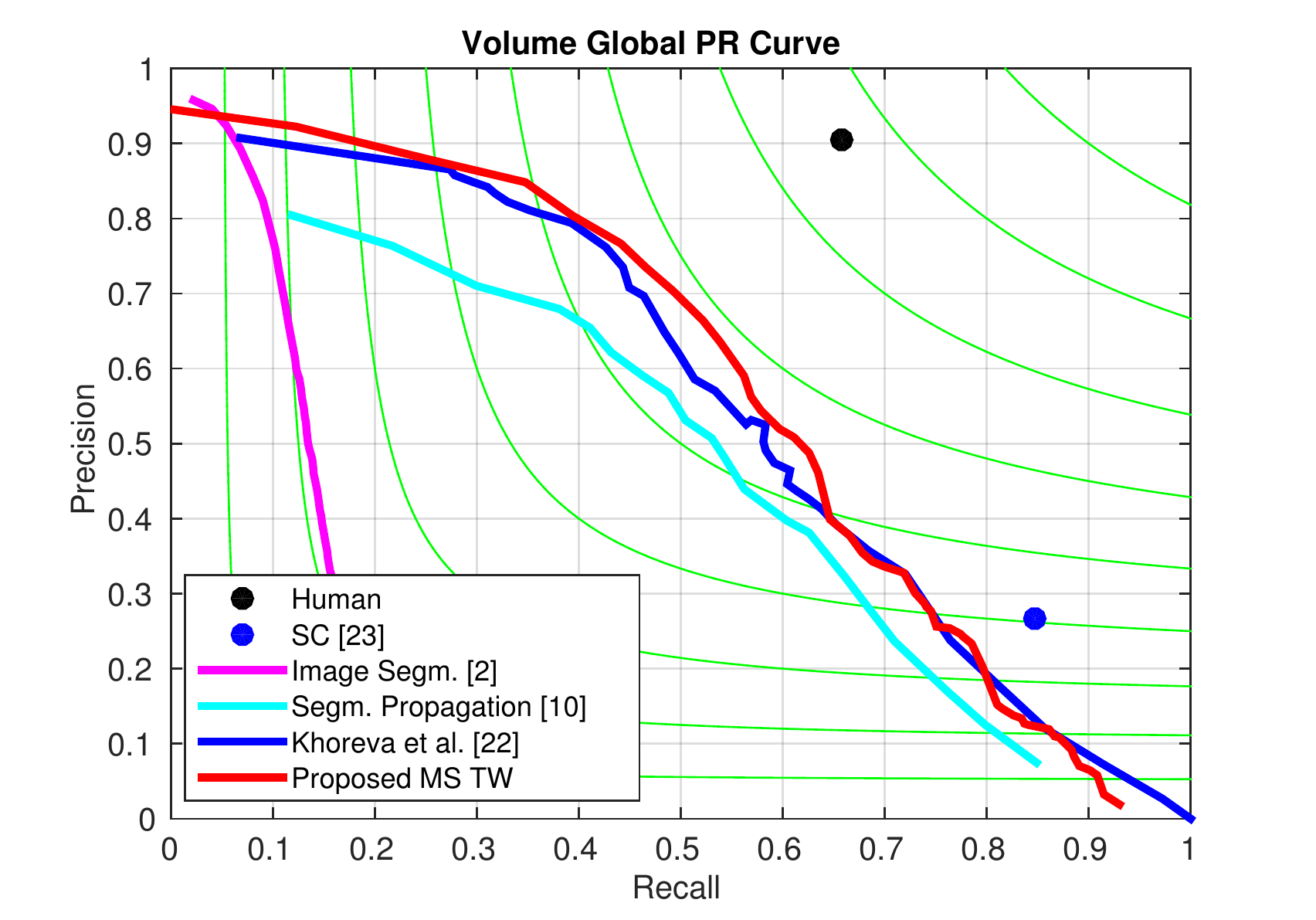}
\caption{Results on the motion subtask the VSB100 \cite{galasso-13,Bro11c} dataset in terms of the region metric VPR.}%However, our curve remains in the low recall regime.}
\label{fig:VSB100_motion}
\end{figure} 
%\paragraph{Spectral Clustering}
%\paragraph{SC + UCM}

\paragraph{Results on Motion Segmentation} 
Since our segmentations claim to be temporally consistent even under significant motion in videos, we also performed the evaluation on the motion subtask of VSB100. For this motion subtask, only a subset of the videos, showing significant motion, is evaluated. Non-moving objects within these videos are not taken into account. Results are reported in Tab. \ref{tab:vs_bvds_motion} and Fig. \ref{fig:VSB100_motion}. As for the general benchmark, our results are outperformed by the state of the art on the BPR but improve over the state of the art in terms of temporal consistency, measured by the region metric VPR.

\section{Conclusions}
We have proposed a method for computing temporally consistent boundaries in videos. To this end, the method builds spatio-temporal affinities based on point-wise mutual information at multiple scales. On the video segmentation benchmark VSB100, the resulting hierarchy of spatio-temporal regions outperforms state-of-the-art methods in terms of temporal consistency as measured by the region metric VPR. We believe that the coarser hierarchy level can help extract high-level content of video. The finer hierarchy levels can serve as temporally consistent spatio-temporal superpixels for learning based video segmentation or action recognition. %We have proposed a method based on the point-wise mutual information measure evaluated at multiple scales for the generation of temporally consistent boundaries in videos. On the video segmentation benchmark VSB100, our proposed hierarchical segmentations outperform state-of-the-art methods in terms of temporal consistency, measured by the region metric VPR. We hope to provide, at different levels in our segmentation hierarchy, both an insight into high-level content of video, and, at a low level in the segmentation hierarchy, a temporally consistent building block that can be used in for higher-level, learning based video segmentation or action recognition methods.
%\paragraph{LMC}
\section*{Acknowledgments}
We acknowledge funding by the ERC Starting Grant VideoLearn.
{\small
\bibliographystyle{ieee}
\bibliography{video_seg}
}

\end{document}